\pdfoutput=1
\documentclass[a4paper, conference]{IEEEtran}

\IEEEoverridecommandlockouts

\usepackage{graphicx}
\usepackage{caption}
\usepackage{longtable}
\usepackage{booktabs}
\usepackage{multirow}
\usepackage{multicol}
\usepackage{tabularx}
\usepackage{textcomp}
\usepackage{ifpdf}
\usepackage[T1]{fontenc}
\usepackage{amssymb}
\usepackage{amsmath}
\usepackage{cite}
\usepackage{enumitem}
\usepackage{eurosym}
\usepackage{listings}
\usepackage{algorithm}
\usepackage{algorithmic}
\usepackage{subcaption}
\usepackage[version=3]{mhchem}
\usepackage{tcolorbox}

\ifCLASSINFOpdf
   \usepackage{epstopdf}
   \graphicspath{{images/}}
   \DeclareGraphicsExtensions{.pdf,.jpeg,.png,.eps}
\else
   \graphicspath{{../images/eps/}}
   \DeclareGraphicsExtensions{.eps}
\fi
\graphicspath{{images/}}

\begin{document}

\title{Towards Energy Impact on AI-Powered 6G IoT Networks: Centralized vs. Decentralized}

\author{\IEEEauthorblockN{Anjie Qiu$^{*}$, Donglin Wang$^{*}$, Sanket Partani$^{*}$,Andreas Weinand$^{*}$ and Hans D. Schotten$^{*+}$}
\IEEEauthorblockA{$^{*}$Institute for Wireless Communication and Navigation (WiCoN), University of Kaiserslautern-Landau (RPTU) \\
$^{+}$Intelligent Networks Research Group (IN), German Research Center for Artificial Intelligence (DFKI)\\
Email: $^{*}$\{qiu, partani, weinand, dwang, schotten\}@eit.uni-kl.de,
$^{+}$\{schotten \}@dfki.de}

}

\maketitle


\begin{abstract}
The emergence of sixth-generation (6G) technologies has introduced new challenges and opportunities for machine learning (ML) applications in Internet of Things (IoT) networks, particularly concerning energy efficiency. As model training and data transmission contribute significantly to energy consumption, optimizing these processes has become critical for sustainable system design. This study first conduct analysis on the energy consumption model for both centralized and decentralized architecture and then presents a testbed deployed within the German railway infrastructure, leveraging sensor data for ML-based predictive maintenance. A comparative analysis of distributed versus Centralized Learning (CL) architectures reveals that distributed models maintain competitive predictive accuracy (\textasciitilde90\%) while reducing overall electricity consumption by up to 70\%. These findings underscore the potential of distributed ML to improve energy efficiency in real-world IoT deployments, particularly by mitigating transmission-related energy costs.
\end{abstract}

\begin{IEEEkeywords}
   6G, Artificial Intelligence, Machine Learning, Energy Saving, Emission Saving, Sustainability
\end{IEEEkeywords}


\section{Introduction}
The integration of Artificial Intelligence (AI) and the Internet of Things (IoT) has advanced applications from smart cities to industrial automation. However, energy consumption remains a critical concern, especially for battery-powered IoT devices. This section compares centralized and decentralized AI-IoT architectures with a focus on energy efficiency.

Centralized architectures rely on a central server for data processing and decision-making, offering uniform control but leading to high energy usage due to data transmission overhead. Transmitting raw data from devices to the cloud increases communication costs, especially in large-scale networks. While optimization techniques exist to reduce energy usage, the centralized model still struggles with scalability and efficiency \cite{zeba2024} \cite{al-abiad2022}. In addition, Edge Computing Integration helps to save energy consumption by processing data at the edge rather than transmitting it all to the cloud. Muhoza et al. \cite{muhoza2023} proposes a Edge-AI enabled IoT architecture with up to 21\% energy saving. Anand et al. \cite{anand2024} proposes an energy-efficient cloud infrastructure for IoT device management to optimize the workload distribution.

\par
However, for energy saving concerns, the decentralized architecture has its advantages against the traditional centralized approach. In decentralized systems, decision-making agents are distributed and processed across IoT devices, eliminating the need for a central server. This approach is more particularly suitable for applications which require low latency, high privacy and robustness against single-point failures. Decentralized systems generally offer better energy efficiency due to reduction of communication overhead, Alenazi et al. \cite{alenazi2021} proposes a distributed ML approach in Cloud Fog Networks resulting in energy savings up to 60\% compared to the centralized one. Zeba et al. \cite{zeba2024} emphasizing optimized workload distribution between edge and cloud resources and enhances system performance and operational efficiency. Also, the decentralized system supports Federated Learning  and Distributed ML architecture, which enable devices to train models locally and updates with model parameters rather than raw data. Applying FL as the fundamental for the distributed architecture, not only preserves privacy but also reduces the energy required for data transmission between data center server and the clients.

\par
This work begins by reviewing recent developments in AI- and ML-driven technologies within the context of 6G IoT systems, followed by an analysis of current energy consumption trends. We propose a quantitative model to evaluate the energy costs of integrating AI and ML into 6G IoT, with a focus on centralized and decentralized architectures. The model aims to clarify the trade-offs between energy efficiency and performance, guiding future sustainable 6G deployments. Empirical validation is carried out using real testbed data from German railway infrastructure, in collaboration with Deutsche Bahn. Results show that decentralized ML significantly reduces energy consumption compared to traditional centralized approaches.

\section{Related Works}
\label{sec:related_worksl}
Training AI models is a resource-intensive process. Chou et al. \cite{chou2024} proposed, that training a single deep neural network can consume thousands of kilowatt-hours (kWh) of electricity, contributing to significant carbon emissions. In the context of IoT, where devices are often resource-constrained, the energy required for training and transmitting data becomes a critical concern. Inference, on the other hand, involves deploying the trained model on edge devices or IoT nodes. While inference typically requires less energy than training, the sheer number of IoT devices and the frequency of data transmission can lead to cumulative energy consumption. For example, Kook et al. \cite{Sujin_Kook_2024} found that transmitting high-dimensional data from IoT devices to edge servers or cloud platforms can result in significant energy overhead.
\par
Data transmission is a major contributor to energy consumption in IoT. Wireless communication, in particular, is energy-intensive due to the need for Radio Frequency (RF) power. Studies of Alkhayyal et al. \cite{Maram_Abdullah_Alkhayyal_2024}) and Shalu et al. \cite{Berin_Shalu_2023}) have shown that transmitting data over long distances or in high-frequency intervals can drastically reduce the battery life of IoT devices. Edge computing, which involves processing data locally on IoT devices or at the edge of the network, has emerged as a promising solution to reduce energy consumption. By minimizing the need for long-distance data transmission, Udayakumar et al.\cite{R_Udayakumar_2023} and Pothakanoori et al. \cite{Kapil_Pothakanoori_2024} both found, that edge computing can significantly lower energy costs compared to traditional cloud computing. 

\par
To address the energy consumption challenges in AI and ML for IoT, researchers have proposed several strategies. Hierarchical Federated Learning (HFL) introduces a layered architecture that combines edge and cloud computing \cite{Chenyu_Gong_2024}. This approach reduces the number of communication rounds required for model convergence, thereby lowering energy consumption. Energy-aware device selection algorithms prioritize IoT devices with higher computational and energy capabilities for participation in FL rounds \cite{Sarah_Kaleem_2023}. This strategy ensures that energy-constrained devices are used efficiently, prolonging their operational lifespan \cite{Alaa_AlZailaa_2023}. Model pruning and quantization techniques reduce the size and computational complexity of AI/ML models. Kook et al.\cite{Sujin_Kook_2022} proposed Energy-efficient data offloading strategies, such as Joint Data Deepening-and-Prefetching (JD2P), reduce the amount of data transmitted by prioritizing feature importance. This approach minimizes energy consumption while maintaining model accuracy.



\section{Methodology}
\label{sec:methodology}

\subsection{Energy Consumption Model}
In this work, we propose a generalized model for energy consumption for the centralized and decentralized IoT system architectures, which provides insights into the carbon emissions and energy consumption, especially focusing on the data transmission and model training. Regarding to our research objective, the learning system we deploy is a simple Convolutional Neural Network (CNN) model, which transforms the input data $x$ into the predictive target $\hat{y}$, where the input dataset $\epsilon_{k}$ are collected from the $K$ devices located within the system. A generally and basic training process of a CNN-based network can be seen in the Algorithm \ref{algo:cnn}: Noted that, in our work the basic Stochastic Gradient Descent (SDG) \cite{sdg_2007} is applied and $Loss(\hat{y}_i, y_i)$ is typically the cross-entropy loss for classification or mean squared error for regression.


\begin{algorithm}[b]
\caption{Supervised Training of a Convolutional Neural Network}
\label{algo:cnn}
\begin{algorithmic}[1]
\REQUIRE Training dataset $\mathcal{D} = \{(x_i, y_i)\}_{i=1}^N$, learning rate $\eta$, number of epochs $E$, batch size $B$
\ENSURE Trained CNN model

\STATE Initialize CNN model parameters $\theta$

\FOR{epoch $= 1$ to $E$}
    \STATE Shuffle training dataset $\mathcal{D}$
    \FOR{each mini-batch $\{(x_j, y_j)\}_{j=1}^B$ from $\mathcal{D}$}
        \STATE $\hat{y}_j \leftarrow \text{CNN}(x_j; \theta)$ \hfill // Forward pass
        \STATE $\mathcal{L} \leftarrow \text{Loss}(\hat{y}_j, y_j)$ \hfill // Compute loss (e.g., cross-entropy)
        \STATE Compute gradients $\nabla_\theta \mathcal{L}$ via backpropagation
        \STATE Update parameters: $\theta \leftarrow \theta - \eta \nabla_\theta \mathcal{L}$
    \ENDFOR
\ENDFOR

\RETURN Trained model with parameters $\theta$
\end{algorithmic}
\end{algorithm}

\par
The energy model mainly consists of two parts, namely the computing and the communication. One life cycle of ML training can be quantized as computing cost $E_{k}^{Compute}$ and communication cost $E_{k,h}^{Trans}$, where the computing cost are usually noted as the energy consumption of data processing, i.e. $E_{k}^{PreP}$, model training $E_{k}^{Train}$. In this work, we quantize the training energy consumption as $b(\epsilon_{k})$ and $b(W)$ bits, where each parameter of the CNN model is fixed number of bits, i.e. 32 bits here. Then, we define the communication energy as the uplink and downlink between the clients and the Base Station (BS) as $E_{k,b}^{(Trans)}$ and $E_{b,k}^{(Trans)}$. In our IoT system scheme, the $K$ devices use the uplink to upload  collected data to the server/backend via the 5G BS, and the server distributed the trained model parameters to the local edge nodes via the downlink communication, if the decentralized approach is applied.

\subsection{Centralized Training: Clients and Server}
In the CL architecture, the model training is usually conducted at the data center or the backend, here $k=S$, where all raw dataset are collected at the sensors and transmitted via the 5G BS. The main training costs are at the server side, including CPU, GPU and other AI-relavent accelerators such as NPU and TPU. Therefore, the generalized energy cost model can be summarized as the Equation \ref{eq:e_cl}:
\begin{equation}
    \centering
    \label{eq:e_cl}
    \begin{split}
        E_{CL}(n) &= E_{CL}^{Compute} + E_{CL}^{Trans} \\
        E_{CL}(n) &= \gamma \cdot n \cdot E_0^{(Compute)} + \alpha \cdot \sum_{k=1}^K b(E_k) \cdot E_{k,0}^{(Trans)}
    \end{split}
\end{equation}

where $E_{CL}^{Compute} = \gamma \cdot n \cdot E_0^{(Compute)}$ represents the computing cost rounding inside the data center with $\gamma \in [0, 1]$ the Power Usage Effectiveness (PUE) of the data center. In \cite{masanet_2013} \cite{capozzoli_2015}, the authors summarized the PUE as the extra electrical power consumed for data storage, power delivery and cooling devices. Andrae et. al. \cite{andrae_2015} conducted researches and estimates on the global electricity usage of communication technology, which including the study on the data centers about how communication cost consists of. Therefore, in order to simply the energy modeling, we apply the PUE with 0.8 in our work for calculating the energy consumption for the data centers, i.e. the computing cost.
\par
Then energy consumed for communication of raw data $E_{CL}^{Trans} = \alpha \cdot \sum_{k=1}^K b(E_k) \cdot E_{k,0}^{(Trans)}$, where $b(\epsilon_{k})$ bits raw data are transmitted for $\alpha$ times of at the $k$-th local training epoch. $\alpha$ represents the times of dataset transmission, since in the reality, data transmission could be failed due complex network condition and re-transmission might be requested. In the simulation, we define the $\alpha=1$ for simplicity, as no transmission failure occurs.

\subsection{Decentralized Training: Edges and Server}
In contrast to CL, the decentralized learning process is usually referred to the FL architecture, where the learning is device- or edge-based and time-varying with synchronized updating \cite{kairouz_2021}. The process of the FL is shown in the Algorithm \ref{algo:fl}, where $W_t$ is the weight matrix of the global model at round $t$, $\mathcal{S}_t$ is the subset of selected clients in round $t$, $W_k^t$ is the weight matrix of local model for client at round $t$, $n_k$ is the size of local dataset on client $k$ and used for weighted averaging.


\begin{algorithm}
\caption{Federated Learning with Multiple Edge Nodes and One Server (FedAvg)}
\label{algo:fl}
\begin{algorithmic}[1]
\REQUIRE Number of communication rounds $T$
\REQUIRE Fraction $C$ of clients selected per round
\REQUIRE Local epochs $E$, Number of clients $K$
\REQUIRE Batch size $B$, Learning rate $\mu$
\ENSURE Trained global model parameters $W_T$

\STATE Initialize global model $W_0$
\FOR{each round $t = 1$ to $T$}
    \STATE Server selects a subset $\mathcal{S}_t \subset \{1, \dots, K\}$ of $m = \max(C \cdot K, 1)$ clients
    \STATE Server sends global model $W_{t-1}$ to all selected clients $k \in \mathcal{S}_t$
    
    \FOR{each client $k \in \mathcal{S}_t$ \textbf{in parallel}}
        \STATE Client $k$ sets $W_k^t \gets W_{t-1}$
        \FOR{epoch $e = 1$ to $E$}
            \STATE Divide local dataset $\mathcal{D}_k$ into batches of size $B$
            \FOR{each batch $(x_i, y_i)$}
                \STATE Compute prediction $\hat{y}_i = f(x_i; W_k^t)$
                \STATE Compute loss $\ell(\hat{y}_i, y_i)$
                \STATE Compute gradients $g \gets \nabla_{W_k^t} \ell$
                \STATE Update model: $W_k^t \gets W_k^t - \mu \cdot g$
            \ENDFOR
        \ENDFOR
        \STATE Client sends updated model $W_k^t$ to server
    \ENDFOR
    
    \STATE Server aggregates models: 
    \[
    W_t \gets \sum_{k \in \mathcal{S}_t} \frac{n_k}{n} W_k^t
    \]
    where $n_k = |\mathcal{D}_k|$, $n = \sum_{k \in \mathcal{S}_t} n_k$
\ENDFOR
\end{algorithmic}
\end{algorithm}

In our work, we deploy the basic aggregation algorithm namely the Federated Averaging (FA) \cite{zhou_2021} to update the global model at the data center. In the decentralized architecture, at the initialization phase, the server distributes the global model to the edge nodes via the 5G BS. Then, the sensors collect raw data and transmitted it to the edge nodes via the Fixed Access Network (FAN), including the Wireless Access Network (WAN) like WiFi or BS \cite{andrae_2015}. After the data collection, edge nodes begins the local training and selected edge nodes can send their model parameters to the server via 5G BS again and update the global model using FA at the data center.Therefore, the energy cost model of the decentralized architecture can be summarized in the Equation \ref{eq:fl}:
\begin{equation}
    \centering
    \label{eq:fl}
    \begin{split}
    E_{FL}(n) &= E_{FL}^{Compute}(n) + E_{FL}^{Trans}(n)
    \end{split}
\end{equation}
Where the computing cost $E_{FL}^{Compute}(n)$ can be calculated as:
\begin{equation}
    \gamma \cdot n \cdot \beta \cdot E_0^{(Compute)} + n \cdot \sum_{k=1}^K E_k^{(Compute)}
    \label{eq:fl_compute}
\end{equation}
and communication cost $E_{FL}^{Trans}(n)$:
\begin{equation}
    b(W) \cdot \left[ n \cdot \sum_{k=1}^K \gamma \cdot E_{0,k}^{(Trans)} + \sum_{t=1}^n \sum_{k \in N_t} E_{k,0}^{(Trans)} \right]
    \label{eq_fl_trans}
\end{equation}

In the configuration of our IoT system, edge nodes are equipped with the embedded low-consumption AI processor and accelerators as the IMX 8 Plus board from NXP \cite{imx8mplus}, we deployed in our work as the edge application platform, where it offers integrated NPUs, energy-efficient architecture, and robust security features, making it a strong choice for deploying ML applications at the edge. The model size $b(W)$ indicates the size in bits of the model parameters to be handled to the server and we output the file as an hdf5, which is an excellent tool for compressing and storing ML model parameters, specially when dealing with large models, distributed training, or Federated Learning. It's reliable, well-supported, and integrates smoothly with ML frameworks.

\subsection{Railway Testbed in Germany by Deutsche Bahn}
This research initiative is underpinned by the German project entitled Second-Level Künstliche Intelligenz in Weichen (SLKI) \cite{slki}, which focuses on developing AI-driven edge applications to enhance digital services for both industrial and public users. A central aim of the project is to explore the energy efficiency of traditional centralized versus decentralized machine learning architectures. In collaboration with Breuer GmbH, Deutsche Bahn has deployed sensors along railway tracks to monitor track conditions and support predictive maintenance. This partnership seeks to enhance data-driven decision-making while reducing energy consumption in railway infrastructure management, ultimately promoting more sustainable practices within the sector.

\begin{figure}[b]
    \centering
    \includegraphics[width=.8\linewidth]{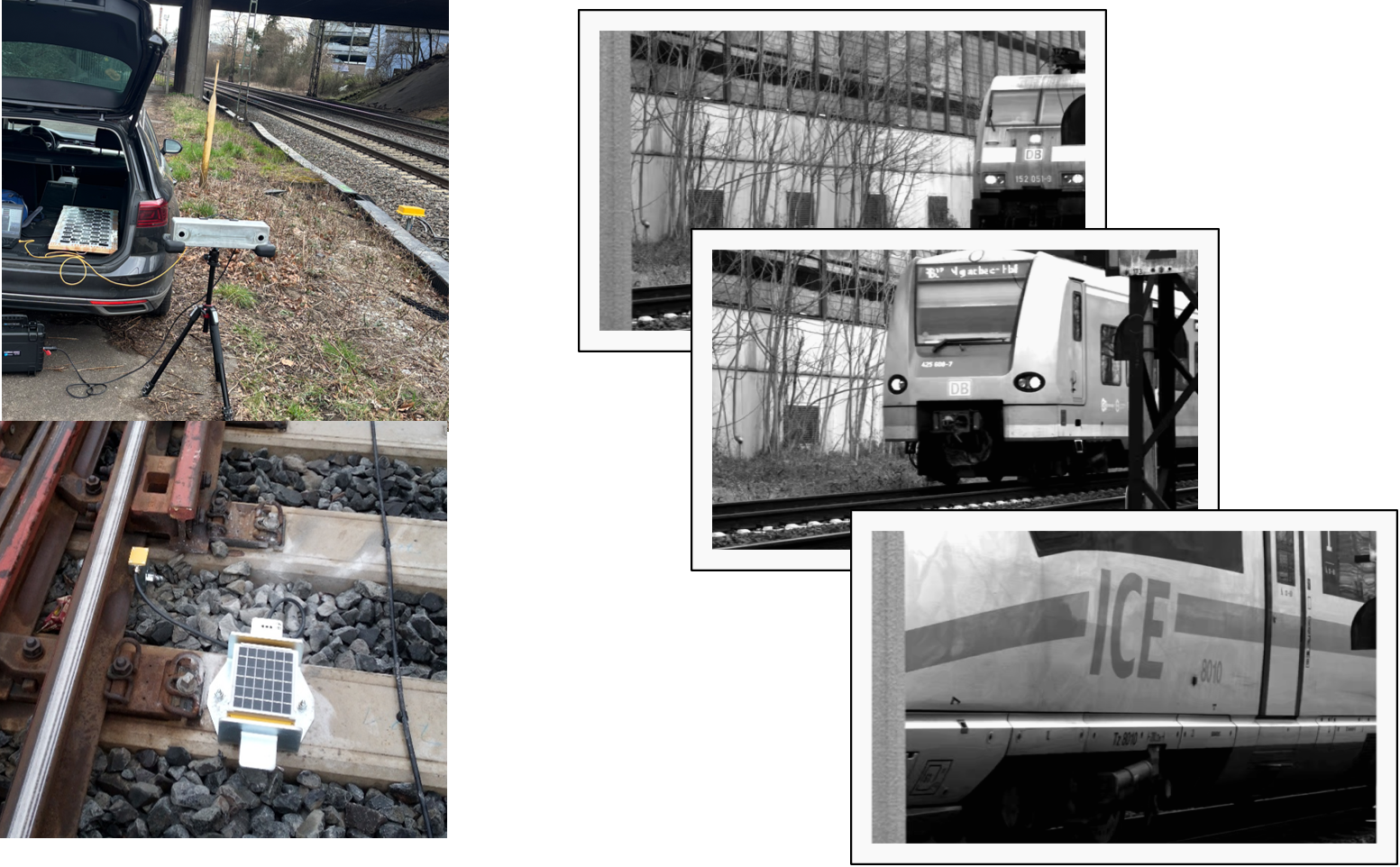}
    \caption{Sensor under the railway track and near the railway with camera, for collecting data and monitoring train speed}
    \label{fig:sensor}
\end{figure}

\begin{figure}[b]
    \centering
    \includegraphics[width=.8\linewidth]{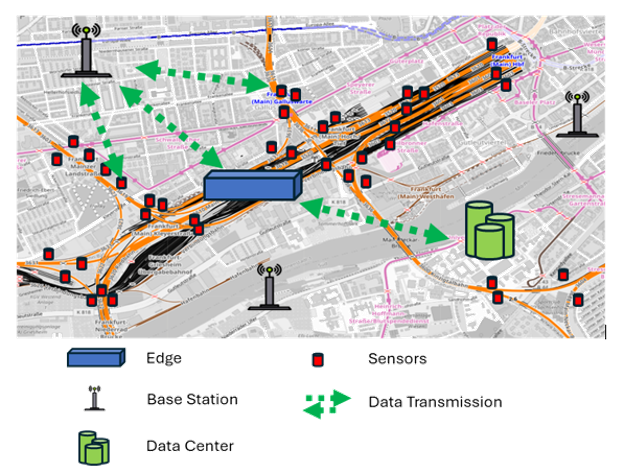}
    \caption{Generalized system-level architecture for the testbed with decentralized learning}
    \label{fig:testbed_fl}
\end{figure}

In our deployment, the centralized approach is first tested and then decentralized approach. In the Figure \ref{fig:testbed_fl}, the generalized IoT system architecture of the decentralized learning is shown, where the red cylinders represents the sensors installed on the railway tracks, transmitting collected raw data to the edge node with blue cuboid via the 5G network indicated by the green dashed arrows. The data center depicted with green cylinders handle the incoming transmission of local model parameters and updating the global model.

\par
Over 1,300 integrated sensor units have been installed across 650 railway turnouts on key rail lines in central and southwestern Germany. These sensors are designed to detect train speeds ranging from 40 to 280 km/h, covering all major train types—long-distance, regional, and freight. This wide-scale deployment significantly enhances data collection capabilities, enabling more accurate predictive maintenance and improved operational efficiency within a 6G IoT framework.

The sensor network operates continuously (24/7) and has been active for nearly a year. During this time, it has generated more than 5 million measurement records, resulting in a dataset exceeding 14.3 terabytes. This extensive dataset provides valuable insights for both predictive maintenance and energy optimization in railway operations. Sensors are installed on each track switch in Figure \ref{fig:sensor}, with additional camera-equipped sensors placed trackside to monitor train speeds.


\section{Result}
\label{sec:result}

\begin{figure}
    \centering
    \includegraphics[width=.8\linewidth]{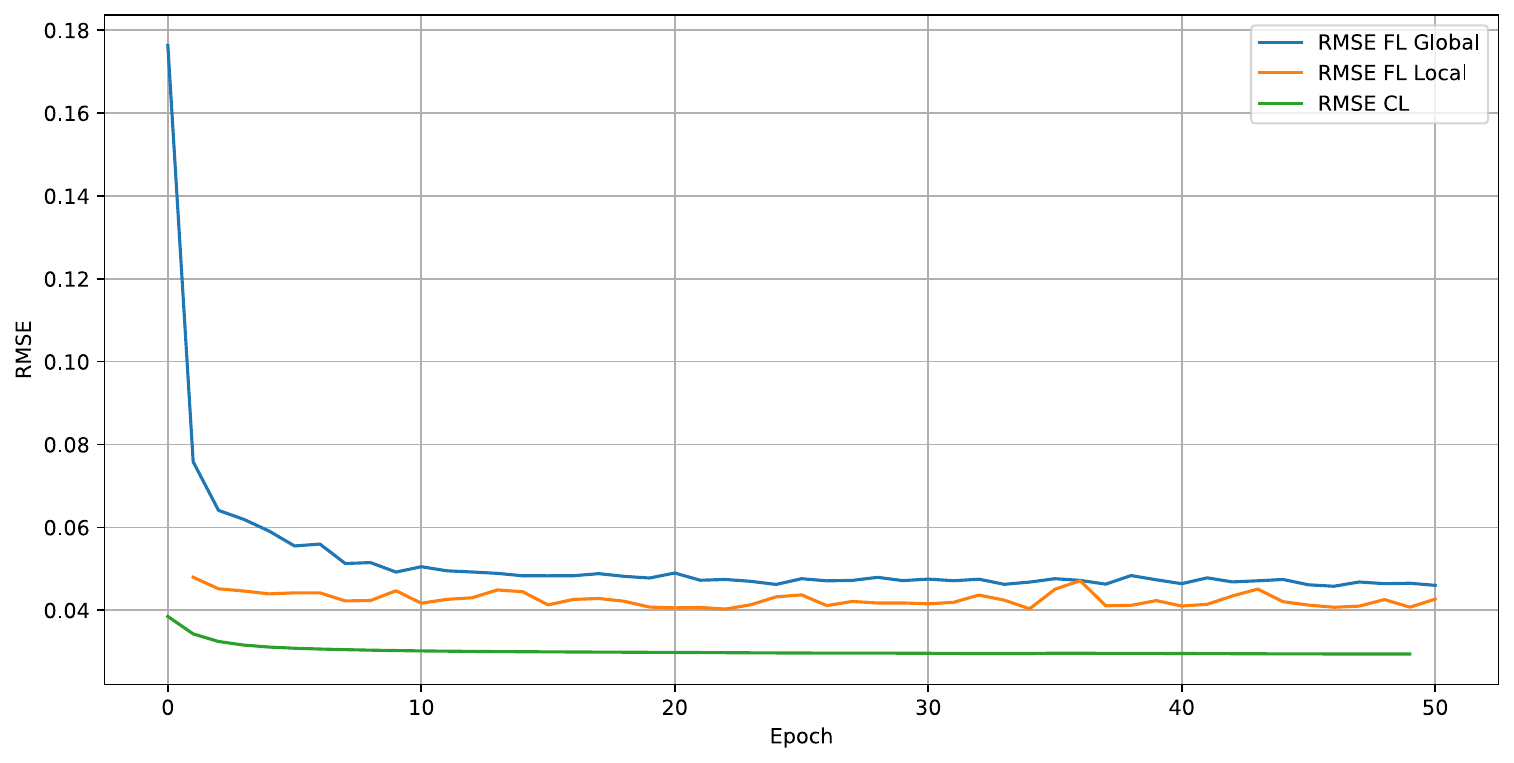}
    \caption{Comparison of prediction performance between CL and FL with RMSE as metric for 50 epochs}
    \label{fig:rmse_cl_fl_train}
\end{figure}

To ensure consistency and objectivity in our comparisons, we first establish a baseline model that reflects the functional objectives and design constraints of the target application. As shown in Figure \ref{fig:rmse_cl_fl_train}, an increase in the number of training epochs typically leads to improved prediction accuracy. However, this is accompanied by a disproportionate rise in energy consumption, driven by the increasing computational complexity of the models. This trend highlights the critical need to balance accuracy and energy efficiency, particularly in ML applications designed for energy-constrained 6G IoT networks. To facilitate this comparison, we normalize the number of training epochs such that both learning paradigms reach similar levels of predictive accuracy. As shown in Figure \ref{fig:rmse_cl_fl_train}, both the centralized and decentralized ML configurations achieve approximately 90\% accuracy after 20 training epochs, equivalent to 20 federate rounds. This convergence enables a fair investigation of their respective energy consumption profiles, providing insight into the scalability, efficiency, and sustainability of AI-driven learning strategies in next-generation IoT systems.

\subsection{Model Training : Performance and Energy Consumption}
\begin{figure}
    \centering
    \includegraphics[width=.8\linewidth]{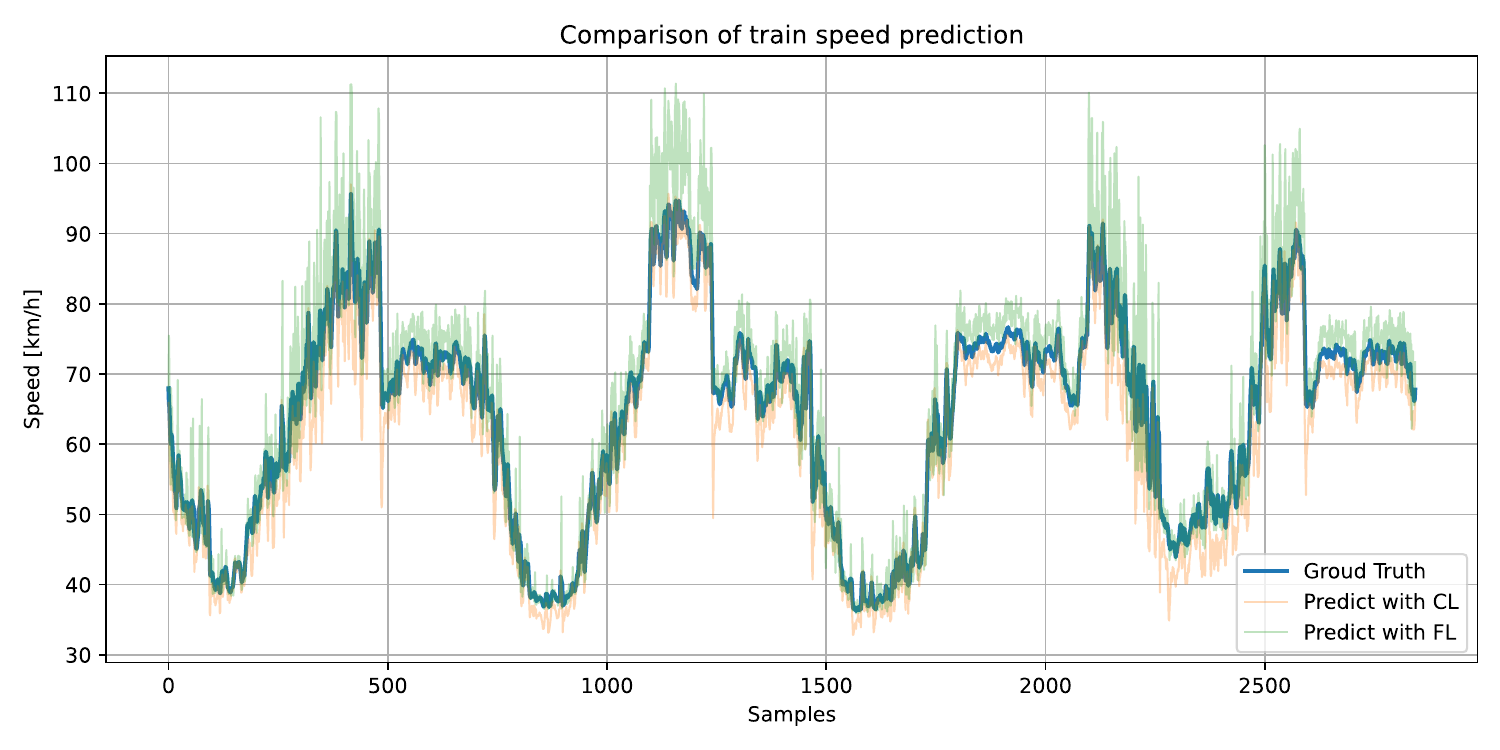}
    \caption{Comparison of train speed prediction with CNN model with centralized and decentralized learning architecture at 20 epochs }
    \label{fig:speed_prediction}
\end{figure}

With the parameter setup mentioned in the previous section, we apply simple CNN model of training the dataset collected from the sensors. And we use the RMSE as the metric of measuring the prediction performance of the CNN model in CL and FL deployment, and the result is shown in Figure \ref{fig:rmse_cl_fl_train}. With the increasing training epochs, the prediction performance is also increased for both CL and FL approach. However, the increment of both deployment convergence quickly after 10 epochs and after 20 epoch the difference is almost not decrease, which indicates that after 20 training epochs the predicting performance of both CL and FL is not increasing with more training rounds .As result, we use the prediction performance over 90\% as the baseline for comparison between centralized and decentralized approaches and the result can be seen in Figure \ref{fig:speed_prediction}. The blue line is the actual measured train speed, and the washed yellow and green lines are the predicted train speed from the CL and FL deployment of our onsite testbed. At 20 training epochs, the prediction performance of the FL is slightly lower as the CL approach, since the computational resource by the CL is more powerful than that in FL, where the edge nodes usually are embedded board with limited computing capability.

\subsection{Data Transmission: Efficiency and Energy Consumption}

After establishing the baseline evaluation criteria for the ML model performance, we proceed to conduct energy measurements for both model training and data transmission. To this end, we implement our measurement campaign across two distinct hardware platforms: one representing a centralized learning architecture, and the other a federated learning setup.

For the centralized architecture, model training is performed on a high-performance desktop computer equipped with ample computational resources, enabling efficient processing of the full raw dataset and training of CNN model. In contrast, the federated architecture relies on the NXP \cite{nxp} i.MX 8M Plus embedded platform \cite{imx8mplus}, which integrates an Arm \cite{arm} Cortex-A53 CPU and a Neural Processing Unit (NPU). For measuring energy consumption, we employ the CodeCarbon  library \cite{codecarbon}, a lightweight and validated Python package capable of estimating electricity usage across computing components including CPU, GPU, NPU and RAM. CodeCarbon also supports estimation of carbon emissions by integrating regional carbon intensity factors based on the geographic location of the device.

\begin{figure}[ht]
  \centering
  \begin{subfigure}[b]{.8\linewidth}
    \includegraphics[width=\linewidth]{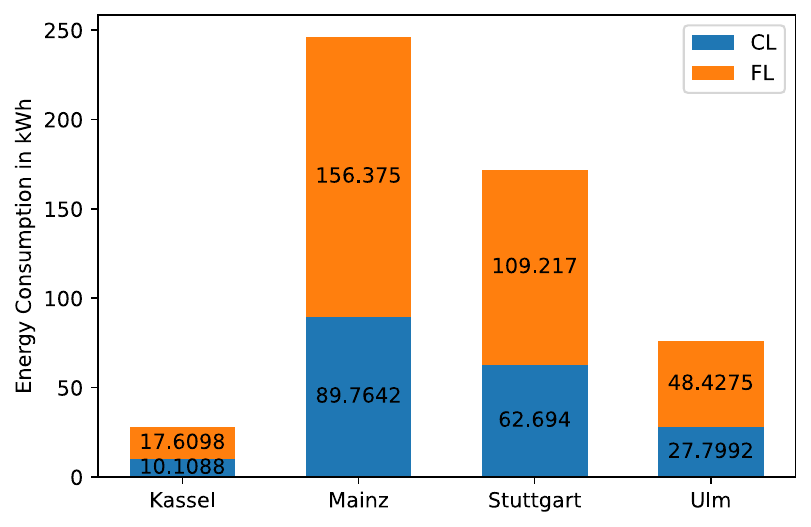}
    \caption{Data transmission}
    \label{fig:energy_cl_fl_breuer:data}
  \end{subfigure}
  \hfill
  \\
  \begin{subfigure}[b]{.8\linewidth}
    \includegraphics[width=\linewidth]{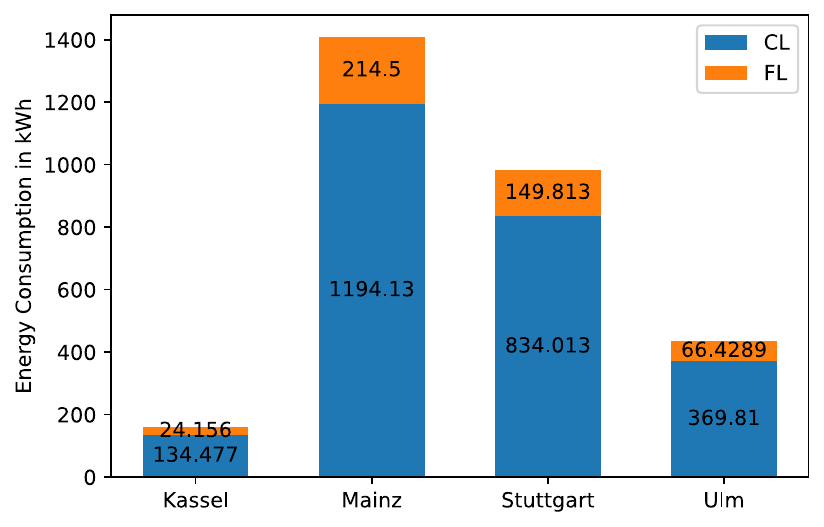}
    \caption{Model training}
    \label{fig:energy_cl_fl_breuer:train}
  \end{subfigure}

  \caption{Comparison of energy consumption for data transmission and model training between CL and FL architecture. a) Data transmission b) Model training}
  \label{fig:energy_cl_fl_breuer}
\end{figure}

The results of the energy consumption analysis for both CL and FL architectures are presented in Figure \ref{fig:energy_cl_fl_breuer}. In Figure \ref{fig:energy_cl_fl_breuer:data}, we observe that the data transmission energy consumption in the FL architecture is higher than in the CL setup. This can be attributed to the increased communication overhead inherent to FL, where bidirectional transmission of local and global model parameters occurs at regular intervals. In contrast, the centralized model only involves one-way upstream data transmission from the sensors to the server, with minimal or no downlink traffic (assuming negligible transmission failures).

Figure \ref{fig:energy_cl_fl_breuer:train} illustrates the energy consumption for model training under both configurations. The results clearly demonstrate that the CL architecture requires substantially more energy for training compared to FL. This is primarily due to the large volume of raw input data processed centrally, which leads to higher computational complexity and, consequently, greater power demands from high-performance hardware. Conversely, the FL setup benefits from lightweight edge devices optimized for low-power operation and designed to handle modest-scale IoT inference and training tasks. Moreover, distributed aggregation of local models in FL can reduce the total training overhead, contributing to additional energy savings.

A summary of the total energy consumption for both learning architectures is provided in Table II, which clearly shows that CL incurs a significantly higher overall energy cost than FL. Furthermore, the results emphasize that data transmission energy is a non-negligible component of the total energy footprint, accounting for up to 70\% of the total energy consumption in the FL configuration. This finding underscores the importance of considering communication efficiency when designing sustainable ML systems for 6G IoT networks.

\begin{table}[b]
    \centering
    \begin{tabular}{c|c|c|c|c}
        Feat. & Kassel & Mainz & Stuttgart & Ulm \\
        \hline
        Number of Sensors  & 208 & 1847 & 1290 & 572 \\
        Energy with CL [kWh] & 144.6 & 1283.9 & 896.7 & 397.6 \\
        Energy with FL [kWh] & 41.7 & 370.8 & 259.0 & 114.8 \\ 
    \end{tabular}
    \caption{Measurement of the total energy consumption for CL and FL on the testbed of German cities}
    \label{tab:energy_total_cl_fl_breuer}
\end{table}


\section{Conclusion}
\label{sec:conclusion}

\subsection{Conclusion}
In conclusion, we have implemented and evaluated a real-world testbed that embodies an AI-driven edge system architecture, serving as a representative model for future 6G-enabled IoT infrastructures. Our deployment utilizes sensors mounted on railway tracks to collect operational data, which is then transmitted via 5G/6G networks to be processed at either edge nodes or a centralized backend server.

A comprehensive comparison was conducted between centralized and decentralized (federated) learning architectures, with particular emphasis on machine learning performance and energy consumption. The experimental results demonstrate that the decentralized ML architecture achieves comparable prediction accuracy to its centralized counterpart while enabling substantial energy savings—up to 70\% reduction in transmission-related energy consumption. This outcome highlights the efficacy and sustainability potential of federated learning approaches for large-scale, energy-constrained IoT systems within the 6G paradigm.

\subsection{Future Scope}
As part of future research, a promising direction involves conducting comparative studies across various IoT application domains, with particular attention to safety-critical systems. In such applications, latency and time-to-result delivery are paramount, raising the open question of whether decentralized (federated) architectures can consistently outperform centralized approaches under stringent real-time constraints.

Another important avenue for exploration is the comprehensive assessment of energy consumption in large-scale AI-integrated IoT deployments. As industry increasingly seeks to embed AI functionalities into nationwide or global-scale IoT systems, it remains unclear whether the energy overhead introduced by AI models is justified by the tangible benefits to industrial productivity and operational efficiency. This raises critical questions regarding the sustainability, scalability, and return on investment of AI-enhanced IoT infrastructures, particularly in the context of national energy strategies and environmental policies.

\section*{Acknowledgment}
This work has been supported by the German Federal Ministry for Transport (BMV) as part of the SLKI project. The authors alone are responsible for the content of the paper.

\bibliographystyle{IEEEtran}
\bibliography{references}

\end{document}